\documentclass{article}

\usepackage{iclr2026/iclr2026_conference,times}

\usepackage{amsmath,amsfonts,bm}

\def\eqref#1{equation~\ref{#1}}

\def\1{\bm{1}}

\DeclareMathAlphabet{\mathsfit}{\encodingdefault}{\sfdefault}{m}{sl}
\SetMathAlphabet{\mathsfit}{bold}{\encodingdefault}{\sfdefault}{bx}{n}

\usepackage{amsmath,amssymb}
\usepackage{enumitem}
\usepackage{hyperref}
\usepackage{url}
\usepackage{booktabs}
\usepackage{graphicx}
\usepackage{placeins}

\title{Execution-Grounded Credit Assignment for GRPO in Code Generation\thanks{Accepted to the ICLR 2026 Workshop on Scaling Post-Training for LLMs (SPOT).}}
\author{
Abhijit Kumar\thanks{Lead author and corresponding author; primary contribution to this work.} \\
\texttt{abhijitkumar4293@gmail.com}
\And
Natalya Kumar \\
\texttt{natalya2kumar@gmail.com}
\And
Shikhar Gupta \\
\texttt{shik1470@gmail.com}
}
\date{}

\iclrfinalcopy

\begin{document}
\maketitle

\begin{abstract}
Critic-free reinforcement learning with verifiable rewards (RLVR) improves code generation by optimizing unit-test pass rates, but GRPO-style updates suffer from coarse credit assignment: a single outcome signal is spread uniformly across long programs even when failure stems from a localized semantic error.

We propose \textbf{Execution-Grounded Credit Assignment (EGCA)}, which localizes GRPO updates using execution traces. For programs that satisfy algorithmic constraints but fail tests, EGCA executes the candidate and a canonical reference solution (curated once offline; used for analysis, not supervision) under identical instrumentation, identifies the earliest semantic divergence, and assigns advantage only to the corresponding token span while masking downstream tokens.

EGCA is a drop-in modification requiring no critic, auxiliary loss, or learned verifier, yielding 82.1\% pass@1 on HumanEval (+3.1 over GRPO) and 68.9\% on MBPP (+1.5) with 18\% wall-clock overhead.
\end{abstract}

\section{Introduction}
Reinforcement learning with verifiable rewards (RLVR), where generated programs are evaluated by unit tests, has become a standard post-training approach for improving code generation models. Critic-free objectives such as GRPO~\citep{shao2024deepseekmath} are appealing: they avoid a value function and directly optimize functional correctness. As base models improve, however, the nature of failures shifts. Modern models increasingly produce code that is syntactically valid, structurally plausible, and fully executable, yet still fails unit tests due to subtle semantic mistakes---an incorrect condition, a misplaced update, or a misinterpreted invariant.

Unit tests provide a reliable correctness signal, but it is temporally coarse: it applies to the entire program rather than to the specific decisions that caused failure. Group-based policy gradients distribute this signal uniformly, so near-correct solutions receive gradients too diffuse to correct localized reasoning errors.

This paper addresses the problem of semantic credit assignment in critic-free RLVR for code generation and targets the near-correct regime where further gains depend on precise attribution rather than coarse feedback.

The following are our key contributions:
\begin{enumerate}[leftmargin=*,itemsep=0.25em]
  \item We show that credit assignment---not reward sparsity---is the main bottleneck in critic-free RL for code generation once models already produce syntactically valid and structurally reasonable programs.
  \item We introduce EGCA, which routes each sample through deterministic failure-mode gates (syntax/constraint/logic) and, for near-correct candidates, localizes the earliest execution divergence against a reference trace to concentrate the GRPO advantage on the causal token span.
  \item We show the approach is agnostic to the debugger's code-generation ability: the trained student surpasses a 1.5B-parameter debugger by +8.2 points, ruling out knowledge distillation as the mechanism.
\end{enumerate}

\begin{figure*}[t]
  \centering
  \includegraphics[width=0.98\textwidth]{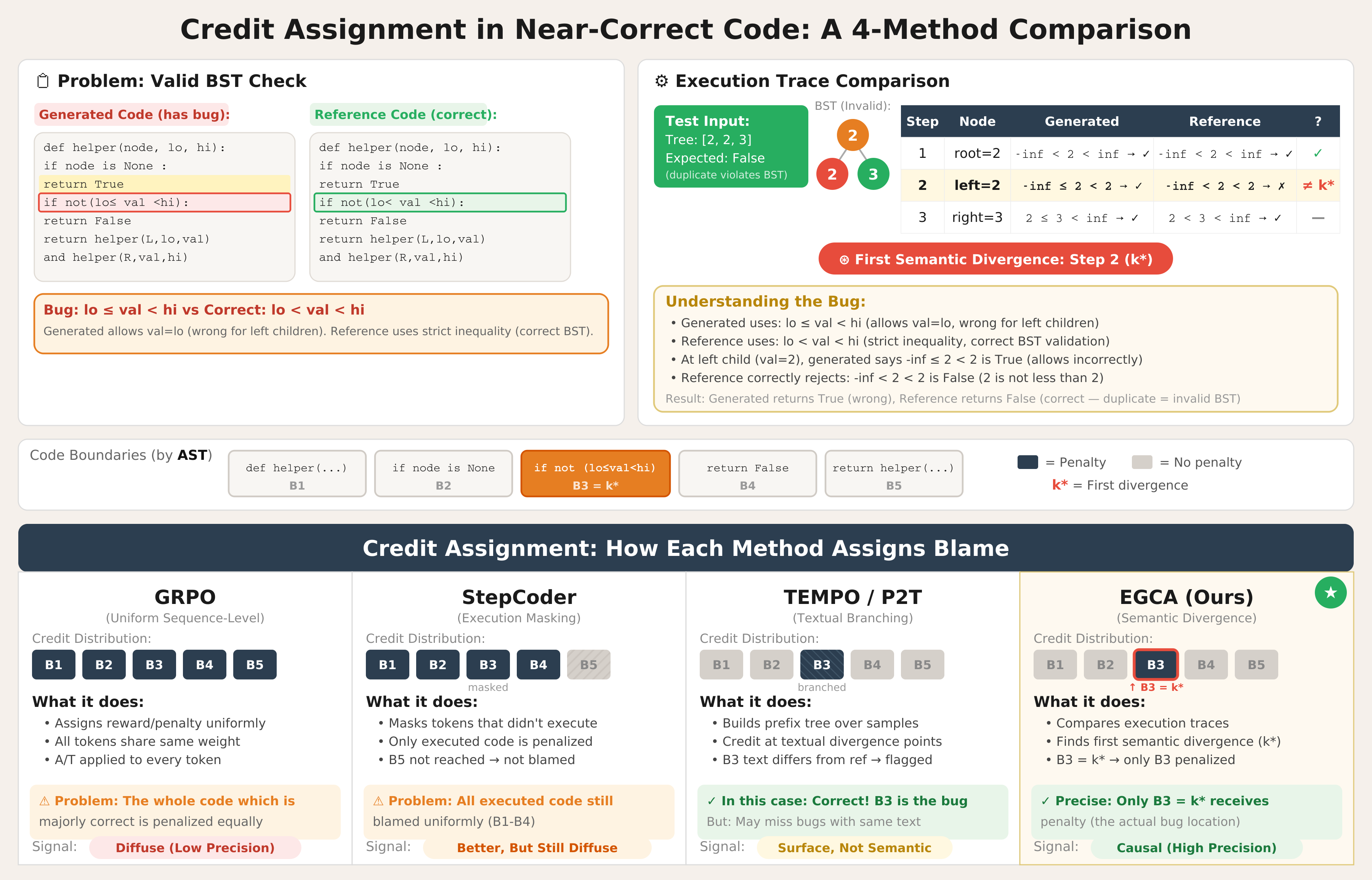}
  \caption{\textbf{Motivation: credit smear vs. localized updates.} Sequence-level RLVR objectives apply unit-test outcomes uniformly across long programs, penalizing large spans of correct code for a localized semantic bug. EGCA concentrates gradient mass on the earliest semantically divergent span (identified via execution) while masking downstream tokens, improving credit assignment in the near-correct regime.}
  \label{fig:credit_assignment}
\end{figure*}

\section{Related Work}
We situate EGCA against five lines of work that densify credit in RLVR for code. The shared limitation is that none reliably localizes failure to semantically causal regions of fully executing programs.

\textbf{Richer outcome signals.} Methods such as RLTF~\citep{liu2023rltf} enrich outcome feedback from execution beyond binary pass/fail, but the signal remains outcome-anchored and does not identify where within a program failure originated.

\textbf{Execution-aware masking.} StepCoder~\citep{dou2024stepcoder} masks unexecuted tokens during updates, reducing spurious blame. However, when programs execute to completion, all tokens are executed, and masking provides no disambiguation among them.

\textbf{Group-structure credit.} Prefix-tree methods such as TEMPO/P2T~\citep{tran2025tempo} derive token-level updates from textual branching points within sample groups. This provides cleaner credit than sequence-level baselines, but textual divergence does not necessarily coincide with the causal location of semantic failure in code.

\textbf{Learned evaluators.} Process reward models~\citep{li2025codeprm,lightman2023verify} train step-level scorers for RL shaping. These meaningfully densify supervision but inherit challenges around label noise and distribution shift from the learned evaluator.

\textbf{Execution semantics and actor--critic.} CodeRL+~\citep{jiang2025coderlplus} adds auxiliary execution-alignment objectives, and actor--critic methods~\citep{le2022coderl,shojaee2023ppocoder} provide dense shaping via learned value functions. Both depart from the critic-free regime.

\textbf{Concurrent directions.} Concurrent work explores complementary directions: RLEF~\citep{gehring2024rlef} grounds code LLMs in execution feedback via RL; multi-turn rewards~\citep{jain2025multiturnrewards} study single-step reward shaping across turns; and MURPHY~\citep{ekbote2025murphy} applies multi-turn GRPO for self-correction. EGCA differs in localizing credit to the earliest divergence within a single generation.

\textbf{Summary and gap.} Existing methods densify feedback or reduce noise, but do not reliably attribute failure to \emph{semantically causal} regions of otherwise well-formed, constraint-following programs. EGCA targets this gap: for near-miss programs that execute but fail due to a localized reasoning error, it identifies the causal span via reference-trace comparison and concentrates the GRPO advantage there.

\section{Method}
\begin{figure*}[t]
  \centering
  \includegraphics[width=0.98\textwidth]{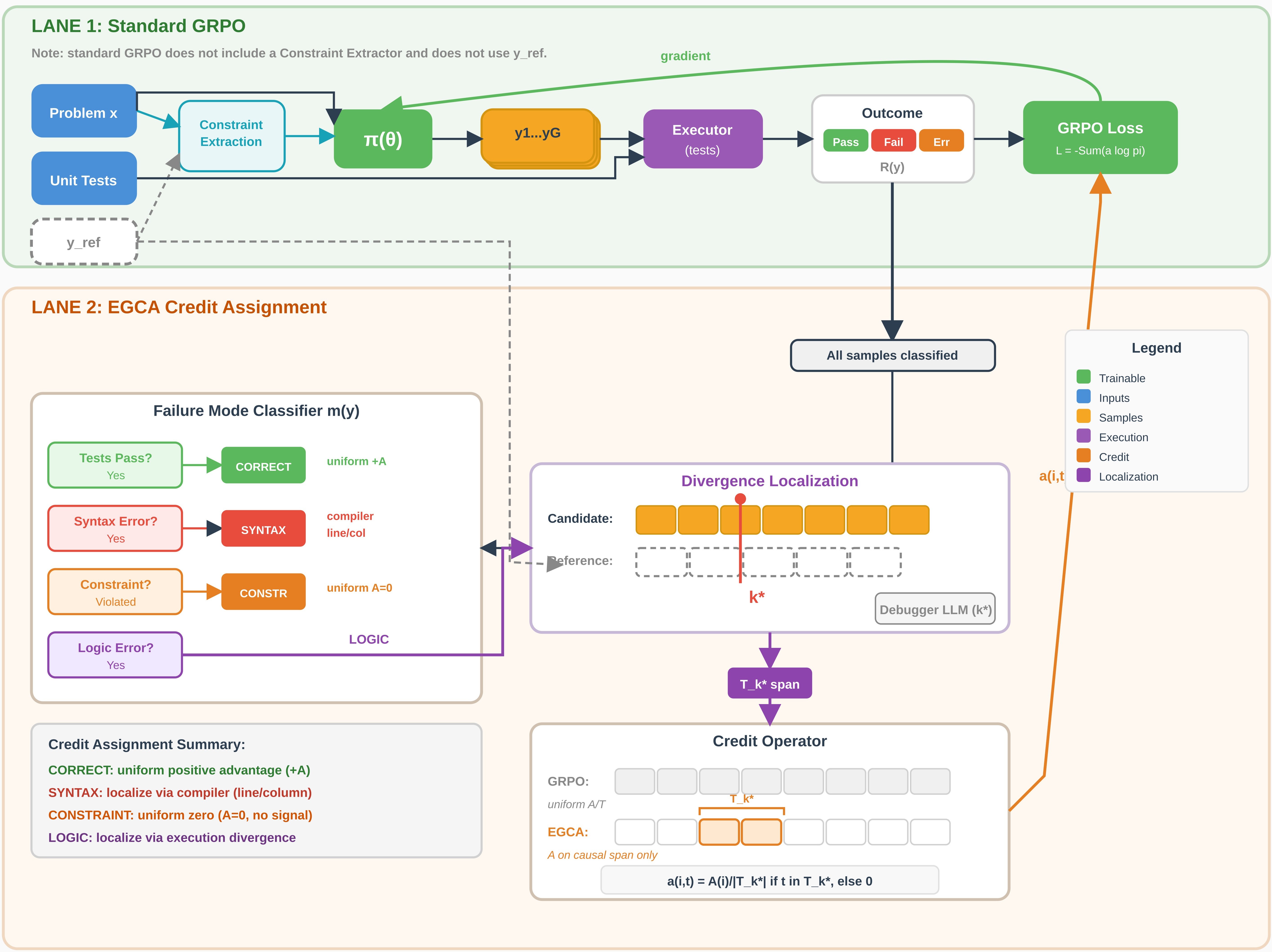}
  \caption{\textbf{EGCA pipeline.} We extract constraints from a canonical reference, sample and execute a group of programs, route each into \textsc{syntax}/\textsc{constraint}/\textsc{logic}/\textsc{correct} via deterministic gates, and apply token-level GRPO by localizing advantage (compiler span for \textsc{syntax}, earliest reference-trace divergence for \textsc{logic}) while masking downstream tokens.}
  \label{fig:pipeline}
\end{figure*}

\subsection{Problem Setting}
Let $x$ denote a programming problem and let $\pi_\theta$ be a code generation policy that samples programs $y\sim\pi_\theta(\cdot\mid x)$. Programs are evaluated using unit tests, producing a base verifiable score $\hat R(y)\in[0,1]$ (fraction of tests passed). We incorporate algorithmic constraints by defining the reward used for optimization as
\begin{equation}
  R(y) = \hat R(y)\,\mathbb{I}_{\mathcal{C}}(y),
\end{equation}
so that a program receives full reward only when it both passes all tests and satisfies the extracted constraints.

We adopt \emph{Group Relative Policy Optimization (GRPO)}~\citep{shao2024deepseekmath}. For each problem $x$, we sample a group of $G$ programs $\{y_i\}_{i=1}^G$ and compute group-relative advantages
\begin{equation}
  A_i = R(y_i) - \frac{1}{G}\sum_{j=1}^G R(y_j).
\end{equation}

\subsection{Canonical Solutions as a Structural Pivot}
For each problem $x$, we assume access to a \emph{canonical reference solution} $y^{\mathrm{ref}}$, curated once offline. The reference is \emph{not} used as a target for imitation. Instead, it serves as a non-parametric pivot for (i) extracting algorithmic constraints, (ii) defining a reference execution behavior, and (iii) anchoring semantic comparisons.
This requirement limits applicability to settings where at least one correct solution exists (e.g., competitive programming and function synthesis with tests); extending EGCA to open-ended generation without references remains future work.

\subsection{Constraint-Guided Sampling}
\paragraph{Constraint extraction.}
A debugging-oriented teacher model extracts from $(x, y^{\mathrm{ref}})$ a set of algorithmic constraints
\begin{equation}
  \mathcal{C} = \{c_1, \dots, c_M\},
\end{equation}
where each constraint specifies a permitted or forbidden structural property (e.g., control-flow form, permitted data structures, or complexity targets). Constraints are non-executable, non-token-level, and solution-agnostic.

\paragraph{Sampling with constraints.}
Constraints are injected as soft guidance via a prompt suffix:
\begin{equation}
  y_i \sim \pi_\theta(\cdot \mid x \;\Vert\; \mathcal{C}).
\end{equation}
This biases sampling toward programs that are structurally comparable to $y^{\mathrm{ref}}$, increasing the density of samples for which semantic divergence is meaningful.

\paragraph{Constraint satisfaction.}
We define a deterministic indicator $\mathbb{I}_{\mathcal{C}}(y)=1$ if $y$ satisfies all constraints in $\mathcal{C}$, and $0$ otherwise.

In addition, semantic divergence is only well-defined when a candidate is \emph{comparable} to the reference at a coarse structural level. We therefore compute a comparability indicator $\mathbb{I}_{\mathrm{cmp}}(y)\in\{0,1\}$ via normalized AST/CFG validation (Section~\ref{sec:structural_alignment}). Candidates that fail this gate are treated as constraint violations in $m(\cdot)$.

\subsection{Structural Validation (Comparability Gate)}\label{sec:structural_alignment}
We parse $y$ and $y^{\mathrm{ref}}$ into ASTs, construct normalized CFGs, compute structural similarity scores, and declare a candidate comparable if scores exceed fixed thresholds, yielding $\mathbb{I}_{\mathrm{cmp}}(y)\in\{0,1\}$.

\subsection{Failure Modes and Credit Operator}
Let $m(y) \in \{\textsc{correct},\textsc{constraint},\textsc{syntax},\textsc{logic}\}$ denote a deterministic failure-mode classifier defined by the following priority order:
\begin{equation}
  m(y)=
  \begin{cases}
    \textsc{syntax} & \text{$y$ raises a compile/runtime error},\\
    \textsc{constraint} & \mathbb{I}_{\mathcal{C}}(y)=0\ \vee\ \mathbb{I}_{\mathrm{cmp}}(y)=0,\\
    \textsc{correct} & \hat R(y)=1\ \wedge\ \mathbb{I}_{\mathcal{C}}(y)=1,\\
    \textsc{logic} & \text{otherwise}.
  \end{cases}
\end{equation}

\noindent (Note that the priority ordering ensures $\mathbb{I}_{\mathrm{cmp}}(y)=1$ for any sample reaching the \textsc{correct} or \textsc{logic} cases.)

We handle syntactic failures before CFG-based validation, since syntax errors can prevent reliable AST/CFG construction; compiler/interpreter diagnostics then provide a precise localization signal for token-level credit assignment.

For a sampled completion $y_i$ of length $T_i$, EGCA defines token-level advantages $a_{i,t}$ as a piecewise function of $m(y_i)$ and a small set of diagnostic spans.

\paragraph{Syntax span.}
If $m(y_i)=\textsc{syntax}$, the compiler/interpreter returns a location; we map it to a token span $\mathcal{T}_{\mathrm{err}} \subset \{1,\dots,T_i\}$ and localize credit to that span, bypassing semantic divergence analysis.\footnote{Our implementation uses Python AST for CFG construction; therefore programs that fail to parse are routed to SYNTAX mode. Extending structural checks with tolerant parsing is possible but not required for the method.}

\paragraph{Divergence span.}
If $m(y_i)=\textsc{logic}$, we localize the earliest semantic divergence against a reference execution to obtain a boundary index $k^*$ and an associated token span $\mathcal{T}_{k^*} \subset \{1,\dots,T_i\}$ (defined below).

\paragraph{Token-level advantage operator.}
We then set
\begin{equation}
  a_{i,t}=
  \begin{cases}
    \displaystyle \frac{A_i}{T_i} & m(y_i)=\textsc{correct},\\
    \displaystyle \frac{A_i}{T_i} & m(y_i)=\textsc{constraint},\\
    \displaystyle \frac{A_i}{|\mathcal{T}_{\mathrm{err}}|}\,\mathbf{1}[t\in\mathcal{T}_{\mathrm{err}}] & m(y_i)=\textsc{syntax},\\
    \displaystyle \frac{A_i}{|\mathcal{T}_{k^*}|}\,\mathbf{1}[t\in\mathcal{T}_{k^*}] & m(y_i)=\textsc{logic}.
  \end{cases}
\end{equation}
The operator is normalized so that $\sum_{t=1}^{T_i} a_{i,t}=A_i$ for every mode, while only localizing credit when blame can be attributed to a specific span.

\subsection{Execution-Grounded Divergence Localization}
We apply divergence localization only when $m(y_i)=\textsc{logic}$, i.e., for candidates that are both constraint-satisfying and comparable to the reference.

\paragraph{Semantic divergence.}
Let $d$ be the first failing unit test input. We define execution boundaries
\begin{equation}
  B(y_i) = (b_1,\dots,b_K),
\end{equation}
each mapping to a token span $\mathcal{T}_k \subset \{1,\dots,T_i\}$. Executing both programs yields state traces
\begin{equation}
  \tau(y_i,d)=(S_1,\dots,S_K),\qquad \tau(y^{\mathrm{ref}},d)=(S^{\mathrm{ref}}_1,\dots,S^{\mathrm{ref}}_K).
\end{equation}
We define the earliest semantic divergence boundary
\begin{equation}
  k^* = \min\{k : S_k \neq S^{\mathrm{ref}}_k\}.
\end{equation}

Because static alignment alone cannot reliably map trace mismatches to fault regions, a debugging-oriented LLM localizes $k^*$ over the aligned structure and paired traces; it is not used as a correctness oracle.

\subsection{Final GRPO Objective}
The overall GRPO objective with token-level advantages is
\begin{equation}
  \mathcal{L}(\theta) = -\sum_{i=1}^{G}\sum_{t=1}^{T_i} a_{i,t}\,\log \pi_\theta\bigl(y_{i,t} \mid x, y_{i,<t}\bigr).
\end{equation}
No teacher gradients, auxiliary losses, or imitation terms are introduced.

\section{Experiments}

\subsection{Benchmarks and Data}
We train our models on APPS+, a curated version of the APPS dataset designed for RL-based code generation. Each problem provides a natural-language specification, a set of executable unit tests, and a canonical reference solution. We focus on Python programs, which dominate APPS+ and allow reliable execution and tracing.

\subsubsection{Canonical Reference Solutions}
Each APPS+ problem includes canonical reference solutions used as described in Section~3.2; they are never used as training targets.

\subsubsection{Evaluation Benchmarks}
We evaluate on HumanEval~\citep{chen2021humaneval} and MBPP~\citep{austin2021mbpp}, reporting pass@1. No benchmark problems appear in training.

\subsection{Training Setup and Implementation Details}
We initialize from DeepSeek-Coder-Instruct-6.7B~\citep{guo2024deepseekcoder} and use Qwen2.5-Coder-7B-Instruct~\citep{hui2024qwen25coder} as the default debugger/localizer. We train with GRPO ($G{=}16$ rollouts, AdamW, lr $=5\times10^{-7}$, $\beta=0.05$, $\varepsilon=0.2$) on APPS+ following StepCoder's RL protocol~\citep{dou2024stepcoder}. Full hyperparameters are in Appendix~H.

\subsection{Baselines}
We compare against representative reinforcement learning--based post-training baselines for code generation, differing primarily in how execution feedback and credit assignment are handled. All baselines use the same backbone model (DeepSeek-Coder-Instruct-6.7B), APPS+ training split, decoding parameters, execution environment, and training budget. The baselines include GRPO with uniform token-level credit assignment, StepCoder, which masks unexecuted code regions during policy updates, RLTF, which leverages multi-granularity unit test feedback without explicit localization of semantic errors, and CodeRL+, which augments reinforcement learning with an auxiliary execution-semantics alignment objective. We follow each method's original training objective and do not introduce constraint guidance, structural validation, or execution-grounded localization into any baseline.

\subsection{Main Results}
\begin{table}[t]
  \centering
  \caption{Main performance (pass@1, \%) on HumanEval and MBPP.}
  \label{tab:main_results}
  \begin{tabular}{lcc}
    \toprule
    Method & HumanEval & MBPP \\
    \midrule
    DeepSeek-Coder (6.7B) base & 78.6 & 65.4 \\
    SFT                         & 71.9 & 60.3 \\
    Vanilla PPO                 & 78.0 & 65.6 \\
    GRPO                        & 79.0 & 67.4 \\
    RLTF                        & 77.9 & 64.5 \\
    StepCoder-mask              & 78.7 & 67.0 \\
    CodeRL+                     & 81.6 & 67.4 \\
    \textbf{EGCA (Ours)}        & \textbf{82.1} & \textbf{68.9} \\
    \bottomrule
  \end{tabular}
\end{table}
Table~\ref{tab:main_results} reports pass@1 on HumanEval and MBPP after training on APPS+ with DeepSeek-Coder-Instruct-6.7B. All methods use identical rollout budgets ($G{=}16$), decoding parameters, and execution environments.

EGCA achieves $82.1\%$ on HumanEval and $68.9\%$ on MBPP. Against vanilla GRPO ($79.0/67.4$), this corresponds to gains of $+3.1$ and $+1.5$ absolute points. Against StepCoder, which masks unexecuted tokens during updates, EGCA improves by $+3.4$ and $+1.9$. This gap suggests that executed-token masking alone is insufficient in the near-correct regime: when programs run to completion, the key question is not which tokens were irrelevant, but which executed decision caused the first semantic deviation.

EGCA also outperforms CodeRL+ ($81.6/67.4$) by $+0.5$ on HumanEval and $+1.5$ on MBPP, despite introducing no auxiliary loss or critic. The improvement derives entirely from reweighting where gradient mass is applied within the standard GRPO objective.

\subsection{Ruling Out Teacher Leakage}
EGCA uses canonical reference solutions and a debugging-oriented LLM to localize divergences, raising a natural concern: do gains stem from distilling teacher knowledge rather than improved credit assignment? We include three controls that suggest the benefit primarily comes from localization.
\begin{table}[t]
  \centering
  \caption{Code-generation capability of debugger LLMs (pass@1, \%).}
  \label{tab:debugger_codegen}
  \begin{tabular}{lcc}
    \toprule
    Model & HumanEval & MBPP \\
    \midrule
    Qwen2.5-Coder-1.5B-Instruct & 70.7 & 69.2 \\
    Qwen2.5-Coder-7B-Instruct   & 84.8 & 79.2 \\
    \bottomrule
  \end{tabular}
\end{table}
\begin{table}[t]
  \centering
  \caption{EGCA performance when varying the debugger/localizer LLM (pass@1, \%).}
  \label{tab:egca_debugger_sweep}
  \begin{tabular}{lcc}
    \toprule
    Debugger model & HumanEval & MBPP \\
    \midrule
    Qwen2.5-Coder-1.5B-Instruct & 78.9 & 66.1 \\
    Qwen2.5-Coder-7B-Instruct   & 82.1 & 68.9 \\
    Claude 4.5 Sonnet           & 83.7 & 67.8 \\
    \bottomrule
  \end{tabular}
\end{table}
\paragraph{Distillation baselines underperform.}
\begin{table}[t]
  \centering
  \caption{Distillation controls (pass@1, \%).}
  \label{tab:distill_controls}
  \begin{tabular}{lcc}
    \toprule
    Method & HumanEval & MBPP \\
    \midrule
    Teacher SFT          & 60.9 & 58.1 \\
    Teacher-critique RL  & 76.3 & 66.1 \\
    \textbf{EGCA (Ours)} & \textbf{82.1} & \textbf{68.9} \\
    \bottomrule
  \end{tabular}
\end{table}

\paragraph{Student exceeds the debugger's generation capability.}
When Qwen2.5-Coder-1.5B-Instruct serves as the debugger, EGCA achieves $78.9\%$ on HumanEval. The same model, used directly as a code generator under the same evaluation protocol, achieves $70.7\%$. The student thus surpasses the debugger by $+8.2$ points, indicating that EGCA extracts a localization signal rather than code-writing competence.

Supervised fine-tuning on teacher-generated code (Teacher SFT) achieves $60.9/58.1$, worse than the base model. Using the teacher as a dense reward signal (Teacher-critique RL) achieves $76.3/66.1$, still $5.8$ points below EGCA on HumanEval. Neither imitation nor teacher-as-judge matches the gains from execution-grounded localization.

\paragraph{Scaling the debugger shows diminishing returns.}
Moving from a $1.5$B to $7$B debugger improves EGCA by $+3.2$ on HumanEval, while a substantially stronger debugger (Sonnet 4.5) adds only $+1.6$ more. If EGCA were primarily distilling solver competence, gains would be expected to track debugger capability; instead, improvements saturate as localization quality stabilizes.

\section{Limitations and Future Work}

EGCA's localization pipeline assumes access to a debugger LLM that can parse execution traces and reason about semantic divergence between two programs.
This model need not generate correct code itself---our experiments confirm that even a 1.5B-parameter debugger whose own pass@1 falls well below the student's still yields meaningful gains---but it must understand code well enough to produce coherent constraints and to compare candidate versus reference traces.
When the debugger lacks this minimum competence, constraint quality degrades, structural comparability judgments become noisy, and the localization signal loses precision.

A related consequence is that EGCA's value is stage-dependent.
The method targets the \emph{near-correct regime}: programs that compile, execute, satisfy structural constraints, and fail only due to localized logical errors.
In our training distribution roughly 35\% of samples fall into this LOGIC mode; the remaining 65\% are handled by uniform or compiler-grounded updates that do not require divergence localization.
As the base policy improves, the fraction of near-correct samples grows and EGCA's advantage compounds.
Conversely, for weak initializations where most failures are syntactic or structural, the localization machinery triggers rarely and the method offers little beyond standard GRPO.
EGCA is therefore most impactful as a later-stage refinement technique applied after the policy already produces broadly reasonable code.

Finally, all experiments here use a 6.7B policy.
Scaling EGCA to larger base models, longer programs, and multi-file generation remains open.
Larger policies will produce more near-correct samples by default, amplifying the regime where EGCA operates, but the debugger LLM and trace infrastructure must scale accordingly.
When multiple structurally distinct solutions are valid, EGCA's constraint extraction and comparability gate may exclude correct but divergent approaches. Correct solutions using structurally divergent approaches may fail the comparability gate; in practice this appears rare in our setting, and such cases could be handled by bypassing constraint checks for programs that pass all tests.

\section{Conclusion}

Credit assignment---not reward sparsity---is the binding constraint on critic-free RL for code generation once models reliably produce executable, structurally sound programs.
EGCA addresses this by converting a coarse unit-test outcome into a token-level policy-gradient operator grounded in runtime semantics: it routes each sample through deterministic failure-mode gates, localizes the earliest execution divergence against a reference trace for near-correct candidates, and concentrates the GRPO advantage on the causal span while masking everything downstream.
The method introduces no critic, no auxiliary loss, and no learned verifier; it changes only which tokens receive gradient mass within the standard objective.

Empirically, EGCA achieves 82.1\% on HumanEval and 68.9\% on MBPP, improving over vanilla GRPO by +3.1 and +1.5 points and over the strongest baseline (CodeRL+) by +0.5 and +1.5, with 18\% wall-clock overhead.
Ablations (Appendix~F) confirm that gains require localizing to the \emph{causally correct} span: random or late-divergence targeting collapses toward the uniform baseline, and softening the mask monotonically erodes performance.
Teacher-leakage controls show the student surpassing its own debugger by over 8 points, ruling out knowledge distillation as the primary mechanism.

The core insight is simple: for near-correct code, knowing \emph{where} a program first goes wrong is more valuable than knowing \emph{that} it goes wrong.
EGCA operationalizes this insight within GRPO, converting execution traces into precise credit without leaving the critic-free regime.

\FloatBarrier
\clearpage
\bibliographystyle{iclr2026/iclr2026_conference}
\bibliography{iclr2026/iclr2026_conference}
\clearpage

\appendix

\section{Additional Details for Reproducibility}

\subsection{Appendix A. Reproducibility Summary}
\paragraph{Compute and schedule.}
We fine-tune using 8$\times$ NVIDIA A100 80GB GPUs with global batch size 64, training for 3 epochs with 0.3-epoch warmup and a linear learning-rate decay to zero.

\subsection{Appendix B. RL Optimization and Generation Settings}
\paragraph{Optimizer hyperparameters.}
During GRPO training, we use a policy learning rate of $5\times 10^{-7}$ with AdamW. Following the critic-free GRPO formulation~\citep{shao2024deepseekmath}, no separate value network is trained.

\paragraph{Note on learning rate.}
Our policy learning rate is more conservative than DeepSeekMath and CodeRL+ (both report $10^{-6}$). We found this smaller learning rate necessary for stable training when combined with token-level credit localization.

\paragraph{Rollouts per prompt.}
For each training example, we sample $G=16$ rollouts using nucleus sampling with temperature $=0.8$, top-$p=0.9$, and maximum output length $=8192$ tokens.

\paragraph{KL regularization and clipping.}
We apply a token-level KL penalty with coefficient $\beta=0.05$ and clip the policy ratio with $\varepsilon=0.2$.

\paragraph{Evaluation-time decoding.}
For inference, we use temperature $=0.2$ and top-$p=0.95$.

\subsection{Appendix C. Execution Environment and Reward Collection}
\paragraph{Runtime.}
Reward collection and evaluations are conducted in a deterministic Python 3.10 sandbox with standard library execution.

\subsection{Appendix D. EGCA Components (Implementation Details)}
This appendix records implementation-specific details for EGCA.

\paragraph{D.1 Execution boundary construction.}
We segment program execution into an ordered sequence of boundary blocks $B=\{B_1,\ldots,B_K\}$, where each $B_k$ corresponds to a contiguous token span and an execution snapshot captured immediately after that span executes.

\paragraph{D.2 Failure-mode routing.}
Each sampled program is assigned to one of four mutually exclusive modes: (i) \textsc{syntax}---parse/compile failure, (ii) \textsc{constraint}---violates extracted algorithmic constraints or fails structural comparability, (iii) \textsc{logic}---executes but fails tests, (iv) \textsc{correct}---passes all tests. Divergence localization is triggered only for \textsc{logic} mode.

\paragraph{D.3 Divergence localization.}
For near-correct samples (\textsc{logic} mode), we identify the earliest boundary index $k^*$ at which the candidate execution diverges from the reference execution under aligned inputs. The localizer LLM (Qwen2.5-Coder-7B-Instruct by default) operates over aligned traces; it is not used as a correctness oracle.

\paragraph{D.4 Token-weighted credit operator.}
Let $A_i$ denote the scalar advantage for sample $i$ under the base GRPO objective. EGCA converts $A_i$ into token-wise advantages $a_{i,t}=w_{i,t}A_i$, where $w_{i,t}$ is concentrated within the localized boundary block and zero elsewhere for \textsc{logic}/\textsc{syntax} modes. Setting all $w_{i,t}=1/T_i$ recovers the uniform baseline.

\subsection{Appendix E. Overhead Accounting}
To make the runtime footprint auditable, we report:
\begin{table}[h]
  \centering
  \caption{Overhead accounting (averages across APPS+ training).}
  \label{tab:overhead}
  \begin{tabular}{lc}
    \toprule
    Metric & Value \\
    \midrule
    Fraction in \textsc{syntax} mode & $\sim 8\%$ \\
    Fraction in \textsc{constraint} mode & $\sim 12\%$ \\
    Fraction in \textsc{logic} (near-correct) mode & $\sim 35\%$ \\
    Fraction in \textsc{correct} mode & $\sim 45\%$ \\
    Average boundaries $K$ per sample & 12.4 \\
    Wall-clock overhead vs. vanilla GRPO & $+18\%$ \\
    Localization trigger rate & $35\%$ of samples \\
    \bottomrule
  \end{tabular}
\end{table}

\subsection{Appendix F. Additional Ablations (Credit Assignment Validity)}
\paragraph{F.1 Localization target.}
\begin{table}[h]
  \centering
  \caption{Ablations on localization target under identical training budgets.}
  \label{tab:ablation_target}
  \begin{tabular}{lcc}
    \toprule
    Target & HumanEval & MBPP \\
    \midrule
    Earliest divergence (EGCA default) & 82.1 & 68.9 \\
    Last divergence & 80.2 & 67.5 \\
    Random executed boundary & 79.6 & 67.1 \\
    No localization (uniform) & 79.0 & 67.4 \\
    \bottomrule
  \end{tabular}
\end{table}

\subsection{Appendix G. Qualitative Case Study: Localized Semantic Errors}
We provide a qualitative analysis showing how execution-grounded localization yields a concentrated learning signal for near-correct programs.

\paragraph{Problem: \texttt{longest\_increasing\_subsequence} (MBPP-style).}
\begin{verbatim}
Write a function that returns the length of the longest strictly increasing
subsequence in a given list of integers.

Example:
>>> longest_increasing_subsequence([10, 9, 2, 5, 3, 7, 101, 18])
4

def longest_increasing_subsequence(nums):
    if not nums:
        return 0

    n = len(nums)
    dp = [1] * n

    for i in range(1, n):
        for j in range(i):
            if nums[i] >= nums[j]:  # BUG: should be strictly greater (>)
                dp[i] = max(dp[i], dp[j] + 1)

    return max(dp)
\end{verbatim}

\paragraph{GRPO output (fails unit tests).}
\begin{verbatim}
def longest_increasing_subsequence(nums):
    if not nums:
        return 0

    n = len(nums)
    dp = [1] * n

    for i in range(1, n):
        for j in range(i):
            if nums[i] >= nums[j]:  # BUG: should be strictly greater (>)
                dp[i] = max(dp[i], dp[j] + 1)

    return max(dp)
\end{verbatim}

\noindent Unit test result: FAILED on input 
\texttt{[1, 3, 3, 5]} (expected 3; got 4).

\paragraph{EGCA output (passes unit tests).}
\begin{verbatim}
def longest_increasing_subsequence(nums):
    if not nums:
        return 0

    n = len(nums)
    dp = [1] * n

    for i in range(1, n):
        for j in range(i):
            if nums[i] > nums[j]:  # CORRECT
                dp[i] = max(dp[i], dp[j] + 1)

    return max(dp)
\end{verbatim}

\paragraph{Execution trace divergence (first semantic mismatch).}
For input \texttt{[1, 3, 3, 5]}, the first semantic divergence occurs at $(i{=}2, j{=}1)$ when comparing equal elements ($3$ vs. $3$).

\begin{table}[h]
  \centering
  \caption{Trace comparison for \texttt{[1,3,3,5]}. The first divergence occurs when the candidate uses $\geq$ while the reference uses $>$.}
  \label{tab:qual_trace}
  \begin{tabular}{cccccc}
    \toprule
    Step & $i$ & $j$ & $\texttt{nums[i]}$ & $\texttt{nums[j]}$ & Divergence? \\
    \midrule
    1 & 1 & 0 & 3 & 1 & No \\
    2 & 2 & 0 & 3 & 1 & No \\
    3 & 2 & 1 & 3 & 3 & Yes \\
    \bottomrule
  \end{tabular}
\end{table}

\paragraph{Credit assignment contrast.}
Under uniform sequence-level GRPO, the negative advantage is spread across all tokens in the program, so the comparison operator receives the same penalty weight as many unrelated tokens. EGCA maps the divergence to the span containing the comparison operator and concentrates the full advantage on that span while masking downstream tokens.

\subsection{Appendix H. Full Hyperparameter Table}
\begin{table}[h]
  \centering
  \caption{Hyperparameters used in our experiments.}
  \label{tab:hyperparams}
  \begin{tabular}{ll}
    \toprule
    Hyperparameter & Value \\
    \midrule
    SFT learning rate & $2\times 10^{-5}$ \\
    SFT epochs & 3 \\
    SFT warmup & 0.3 epochs \\
    SFT LR schedule & Linear decay to zero \\
    \midrule
    GRPO policy learning rate & $5\times 10^{-7}$ \\
    GRPO optimizer & AdamW \\
    Rollouts per prompt ($G$) & 16 \\
    Train sampling temperature & 0.8 \\
    Train sampling top-$p$ & 0.9 \\
    Max generation tokens & 8192 \\
    KL coefficient ($\beta$) & 0.05 \\
    Clip epsilon ($\varepsilon$) & 0.2 \\
    \midrule
    Eval decoding temperature & 0.2 \\
    Eval decoding top-$p$ & 0.95 \\
    \midrule
    Hardware & 8$\times$ A100 80GB \\
    Global batch size & 64 \\
    \bottomrule
  \end{tabular}
\end{table}

\end{document}